# Translating the Grievance Dictionary: a psychometric evaluation of Dutch, German, and Italian versions

Isabelle van der Vegt[1], Bennett Kleinberg[2,3], Marilu Miotto[4], Jonas Festor[2]

Correspondence: i.w.j.vandervegt@uu.nl

[1]Department of Sociology, Utrecht University, [2]Department of Methodology and Statistics, Tilburg University, [3]Department of Security and Crime Science, University College London, [4]Erasmus School of Social and Behavioural Sciences, Erasmus University

## Abstract

This paper introduces and evaluates three translations of the Grievance Dictionary, a psycholinguistic dictionary for the analysis of violent, threatening or grievance-fuelled texts. Considering the relevance of these themes in languages beyond English, we translated the Grievance Dictionary to Dutch, German, and Italian. We describe the process of automated translation supplemented by human annotation. Psychometric analyses are performed, including internal reliability of dictionary categories and correlations with the LIWC dictionary. The Dutch and German translations perform similarly to the original English version, whereas the Italian dictionary shows low reliability for some categories. Finally, we make suggestions for further validation and application of the dictionary, as well as for future dictionary translations following a similar approach.

**Background**

The Grievance dictionary was developed for the purpose of automatically analysing language use in violent, threatening or grievance-fuelled texts (van der Vegt et al., 2021). It was developed by consulting threat assessment professionals, who assess the threat of violence posed by a person. Sometimes, this assessment is based on written communication, such as threat letters or social media posts, and thus made based on linguistic information. The Grievance Dictionary could aid in this process by measuring concepts relevant to threat assessment, thereby reducing the time spent on manual assessment, especially when large amounts of text data need to be examined.

The original paper (van der Vegt et al., 2021) describes the process of determining dictionary categories through consultation, word list annotation through crowdsourcing, psychometric evaluation, and various validation tests. To date, the dictionary has also been utilized in a wide variety of academic papers using textual data in English[1]. Several have applied the dictionary in the context of extremism, such as Allchorn et al. (2022) who used the dictionary to compare violent and non-violent violent extremist manifestos, Ai et al. (2024) who applied it to QAnon conspiracy videos, and Habib et al. (2022) who analysed language on radical misogynist subreddits. The dictionary has also been applied in other domains, for example to compare language use on FOX to MSNBC (Lanning et al., 2021) and to study verbal aggression in the German parliament (Klonek et al., 2022). In the latter study, German speeches were translated to English in order to apply the dictionary, demonstrating the possible need for translations of the Grievance dictionary.

Given the broad range of themes described above and the fact that threats of violence are not limited to the English language, we have translated the Grievance Dictionary to other languages, in order to enhance its applicability. Earlier work translating the LIWC dictionary from English to Dutch has shown that automatic translation with subsequent human correction performs nearly as well as a fully human translated version (Boot et al., 2017). In our approach, we similarly opt for automated translations corrected by two human annotators. The current paper details the translation process into Dutch, German, and Italian. Psychometric characteristics and correlations with the general purpose dictionary LIWC (Pennebaker et al., 2015) are reported for each language.

**Method**

*Data and code availability*
The code for reproducing the analysis is available via the Open Science Framework: https://osf.io/3grd6/. Dictionary translations (and code to run them) are available via the Github repository: https://github.com/Isabellevdv/grievancedictionary. For further information on dictionary usage and additional materials, please refer to the website: https://grievancedictionary.net/

*Translation and agreement*
All Grievance dictionary terms with an average goodness-of-fit rating of at least 7 were translated ($n$=5,513 terms). The original, lowercased, unstemmed terms were used for translation purposes (the stemmed dictionary contains $n$=3,633 terms). The googleLanguageR package (Edmondson et al., 2020) was used to translate the English terms into German, Dutch, and Italian via the Google Translate API (see Table 1 for examples). All translations were subsequently annotated by a native speaker of each language, following the below criteria:
1. Is the translation semantically correct?
2. Is the translation correct given the context (i.e., dictionary category)?
3. Should the word be deleted or replaced by a better alternative translation?
4. Are there additional – equally appropriate – translations?

---

[1] See https://grievancedictionary.net/ for an overview of papers using the dictionary.

*Table 1.* Example translations

| Category | English | Dutch | German | Italian |
|---|---|---|---|---|
| desperation | last resort | laatste redmiddel | letzter Ausweg | ultima risorsa |
| frustration | irritated | geïrriteerd | irritiert | irritata |
| god | prayer | gebed | Gebet | preghiera |
| weaponry | ammunition | munitie | Munition | munizioni |

To assess reliability of annotations, for each language, a second native speaker assessed a random sample of translations along the same criteria. 25 terms from each of the 22 categories were randomly sampled resulting in 550 translations per language to be annotated. This corresponds to 10% of all translated terms per language. Agreement between the first and second annotators was assessed on whether the translation was semantically correct and contextually appropriate (i.e., criteria 1 and 2). Annotators were considered to be in agreement if they both stated that a term should (not) be corrected. That is, agreement was calculated based on binary decisions by the annotators (correctly translated or not correctly translated). Given that annotations were highly skewed toward being correctly translated, we opted for Gwet's AC1 statistic for inter-rater reliability. This procedure is preferred over Cohen's Kappa when there is a strong imbalance between marginal totals, as is the case here (Gwet, 2008; Wongpakaran et al., 2013). Gwet's AC1 ranges from -1 (poor agreement) to 1 (perfect agreement).

*Psychometric analysis*
The psychometric analysis followed the original procedure from the English version of the Grievance dictionary. First, the internal reliability of the dictionary categories was computed by means of Cronbach's alpha scores. Specifically, it was tested whether words in each category yielded similar scores for the category (i.e., similar to computing reliability for a survey scale), using the proportional occurrence of each word in the 22 categories. The average reliability was computed across two different corpora for each language, see Table 2. Second, Pearson correlations were computed between the Grievance Dictionary categories and the LIWC dictionary using the same corpora (Table 2). For this analysis, Dutch (Boot et al., 2017) and German (Meier et al., 2019) versions of the LIWC 2015 were utilized. In Italian, only a translation of the 2007 LIWC was available[2]. We adjusted for multiple hypothesis testing using a Bonferroni correction ($p<0.05/22$ categories).

*Table 2.* Corpora for analysis

| Language | Corpus | Number of documents (number of tokens) |
|---|---|---|
| Dutch | Dutch Book Reviews Dataset (Burgh & Verberne, 2019) | 1,112 (277,030) |
|  | Personae Corpus (Luyckx & Daelemans, 2008) | 145 (205,686) |
| German | German Online Discussions (Schabus et al., 2017) | 10,274 (4,291,084) |
|  | German Abusive Comments (Assenmacher et al., 2021) | 57,405 (2,339,248) |
| Italian | FEEL-IT Twitter posts (Bianchi et al., 2021) | 2,037 (52,443) |
|  | Reddit Italy "Caffe Italia" (Cerone, 2023) | 318,214 (10,143,498) |

**Results**

*Agreement between annotators*
Across translations, we see high levels of agreement between annotators, with perfect agreement for some categories. Results are shown below in Table 3. Given the high level of agreement between the two annotators per language, we opted to incorporate all suggestions from the first annotator into the final dictionary. This means that translations from Google Translate deemed incorrect by the first annotator were removed or replaced with the suggestions of this annotator (criteria 1 and 2). Any additional translations suggested by the first annotator were added to the dictionary (criteria 3 and 4). See Table 4 below for an overview of this process. After stemming all words, this procedure resulted in final dictionaries of 4,326 words in Dutch, 4,629 words in German, and 4,171 words in Italian (compared to 3,663 for the stemmed English version).

---
[2] To our knowledge, there is no academic paper associated with the Italian 2007 translation. We used the version available in the dictionary repository for LIWC-22 provided by dr. A. Rellini.

*Table 3.* Agreement between annotators

| Category | Dutch | | German | | Italian | |
|---|---|---|---|---|---|---|
| | No. terms | AC1 [CI] | No. terms | AC1 [CI] | No. terms | AC1 [CI] |
| Deadline | 187 | 0.63 [0.32-0.95] | 217 | 0.71 [0.19 - 1.00] | 164 | 0.81 [0.60 - 1.00] |
| Desperation | 152 | 0.86 [0.68 – 1.00] | 175 | 0.60 [0.00 - 1.00] | 159 | 0.91 [0.78 - 1.00] |
| Fixation | 85 | 0.81 [0.60 – 1.00] | 81 | 0.76 [0.32 - 1.00] | 75 | 0.91 [0.76 - 1.00] |
| Frustration | 126 | 0.90 [0.75 – 1.00] | 137 | 1.00 | 130 | 0.91 [0.78 - 1.00] |
| God | 233 | 0.74 [0.47 – 1.00] | 247 | 0.76 [0.32 - 1.00] | 227 | 0.91 [0.78 - 1.00] |
| Grievance | 92 | 1.00 | 93 | 0.76 [0.32 - 1.00] | 96 | 0.96 [0.87 - 1.00] |
| Hate | 186 | 0.95 [0.86 – 1.00] | 199 | 1.00 | 194 | 0.91 [0.76 - 1.00] |
| Help | 164 | 0.80 [0.57 - 1.00] | 171 | 0.76 [0.32 - 1.00] | 153 | 0.86 [0.69 - 1.00] |
| Honour | 141 | 0.85 [0.67 - 1.00] | 147 | 1.00 | 131 | 1.00 |
| Impostor | 141 | 0.81 [0.60 - 1.00] | 142 | 0.76 [0.32 - 1.00] | 130 | 0.85 [0.67 - 1.00] |
| Jealousy | 109 | 0.86 [0.69 - 1.00] | 112 | 0.89 [0.62 - 1.00] | 107 | 1.00 |
| Loneliness | 114 | 0.74 [0.47 - 1.00] | 118 | 0.89 [0.62 - 1.00] | 110 | 0.80 [0.57 - 1.00] |
| Murder | 356 | 0.72 [0.45 - 1.00] | 369 | 0.89 [0.62 - 1.00] | 328 | 0.80 [0.57 - 1.00] |
| Paranoia | 164 | 0.85 [0.67 - 1.00] | 165 | 0.89 [0.62 - 1.00] | 148 | 0.91 [0.78 - 1.00] |
| Planning | 250 | 0.83 [0.62 - 1.00] | 261 | 0.89 [0.62 - 1.00] | 232 | 0.91 [0.76 - 1.00] |
| Relationship | 328 | 0.86 [0.69 - 1.00] | 357 | 0.89 [0.62 - 1.00] | 307 | 0.86 [0.69 - 1.00] |
| Soldier | 311 | 0.85 [0.66 - 1.00] | 341 | 0.76 [0.32 - 1.00] | 325 | 0.96 [0.87 - 1.00] |
| Suicide | 209 | 0.79 [0.55 - 1.00] | 221 | 0.89 [0.62 - 1.00] | 198 | 0.91 [0.76 - 1.00] |
| Surveillance | 261 | 0.87 [0.67 - 1.00] | 278 | 0.89 [0.62 - 1.00] | 244 | 0.91 [0.78 - 1.00] |
| Threat | 180 | 0.96 [0.87 - 1.00] | 201 | 1.00 | 171 | 0.91 [0.78 - 1.00] |
| Violence | 302 | 0.89 [0.71 - 1.00] | 323 | 1.00 | 290 | 0.96 [0.87 - 1.00] |
| Weaponry | 235 | 0.93 [0.77 - 1.00] | 274 | 1.00 | 252 | 0.96 [0.87 - 1.00] |
| Total | 4,326 | | 4,629 | | 4,171 | |

*Table 4.* Corrections made in the translation evaluation process per language

| Language | Correctly translated | Words corrected | New words | Unstemmed dictionary | Stemmed dictionary (final) |
|---|---|---|---|---|---|
| Dutch | 5,023 | 327 | 98 | 5,448 | 4,326 |
| German | 5,447 | 66 | 149 | 5,662 | 4,629 |
| Italian | 5,160 | 353 | 97 | 5,610 | 4,171 |

*Internal reliability*

The average Cronbach's alpha scores across corpora for each table are reported in Table 5. In Dutch, the highest reliability is found for the category 'violence' (0.43), whereas the lowest score of 0.11 is found for 'help'. The range is somewhat similar to that achieved in the English version (0.12-0.37). In German, the scores are somewhat lower, ranging between 0.05 and 0.27. In Italian, the alpha scores are substantially lower, with a maximum of 0.14 achieved for 'murder'. In one case (fixation), there appears to be a negative covariation between words in the category.

*Table 5.* Internal reliability (Cronbach's alpha) per Grievance category and translation

| Category | Dutch | German | Italian | English (original) |
|---|---|---|---|---|
| Deadline | 0.22 | 0.05 | 0.06 | 0.27 |
| Desperation | 0.15 | 0.09 | 0.04 | 0.21 |
| Fixation | 0.19 | 0.05 | -0.01 | 0.12 |
| Frustration | 0.24 | 0.07 | 0.01 | 0.22 |
| God | 0.31 | 0.20 | 0.09 | 0.35 |
| Grievance | 0.18 | 0.08 | 0.02 | 0.16 |
| Hate | 0.21 | 0.16 | 0.09 | 0.30 |
| Help | 0.11 | 0.06 | 0.03 | 0.19 |
| Honour | 0.17 | 0.06 | 0.04 | 0.26 |
| Impostor | 0.17 | 0.09 | 0.06 | 0.19 |
| Jealousy | 0.20 | 0.05 | 0.01 | 0.21 |
| Loneliness | 0.15 | 0.05 | 0.00 | 0.18 |
| Murder | 0.32 | 0.24 | 0.14 | 0.35 |
| Paranoia | 0.20 | 0.10 | 0.06 | 0.23 |
| Planning | 0.19 | 0.11 | 0.06 | 0.31 |
| Relationship | 0.35 | 0.17 | 0.06 | 0.33 |
| Soldier | 0.29 | 0.26 | 0.09 | 0.37 |
| Suicide | 0.18 | 0.16 | 0.05 | 0.26 |
| Surveillance | 0.23 | 0.14 | 0.09 | 0.25 |
| Threat | 0.33 | 0.20 | 0.07 | 0.30 |
| Violence | 0.43 | 0.24 | 0.12 | 0.36 |
| Weaponry | 0.27 | 0.27 | 0.09 | 0.34 |

*Correlations with LIWC*

Below, the three highest correlations between LIWC dictionary categories and Grievance Dictionary categories in Dutch, German, and Italian are shown (Table 6, 7, and 8, respectively). All reported correlations were statistically significant at the $p<0.0023$ (0.05/22 categories) level. Overall, correlations are small. In Dutch, correlations range between $r=0.07$ to $r=0.25$. For some Grievance categories (e.g., frustration, suicide), only 1-2 statistically significant correlations with LIWC variables were found. Similar to the original English language version, correlations were found between Grievance dictionary and LIWC categories that could be argued to be psychologically related, such as between 'deadline' and 'focus on the future', and 'threat' and 'death'. In German, correlations are small to medium ($r=0.02$ to $r=0.44$). Again, psychologically related variables are correlated here, such as between 'grievance' and 'negative emotion' and between 'planning' and 'drives'. In Italian, correlations range between $r=0.04$ to $r=0.55$. Examples of psychologically meaningful correlations are between 'frustration' and 'negative emotion', and 'murder' with 'death'.

*Table 6.* Top 3 Strongest Pearson Correlations Dutch dictionary and LIWC

| Category | Strongest correlating categories | | |
|---|---|---|---|
| Deadline | time: 0.19 [0.1-0.28] | discrep: 0.10 [0.06-0.14] | focusfut: 0.08 [0.04-0.13] |
| Desperation | sad: 0.13 [0.06-0.2] | Dic: 0.11 [0.07-0.15] | Exclam: 0.10 [0.06-0.14] |
| Fixation | focuspres: 0.07 [0.03-0.12] | ppron: 0.07 [0.03-0.12] | NS[1] |
| Frustration | feel: 0.09 [0.05-0.13] | NS | NS |
| God | see: 0.16 [0.12-0.2] | percept: 0.14 [0.10-0.18] | male: 0.11 [0.07-0.15] |
| Grievance | social: 0.11 [0.05-0.17] | anx: 0.09 [0.05-0.13] | male: 0.09 [0.05-0.13] |
| Hate | shehe: 0.12 [0.08-0.16] | see: 0.10 [0.06-0.14] | male: 0.09 [0.05-0.13] |
| Help | posemo: 0.17 [0.08-0.25] | affect: 0.16 [0.07-0.26] | achieve: 0.15 [0.06-0.24] |
| Impostor | anger: 0.21 [0.16-0.26] | ipron: 0.12 [0.08-0.16] | NS |
| Jealousy | negemo: 0.13 [0.09-0.17] | social: 0.10 [0.06-0.15] | ipron: 0.09 [0.05-0.13] |
| Loneliness | affect: 0.13 [0.05-0.21] | NS | NS |
| Murder | negemo: 0.18 [0.09-0.28] | insight: 0.17 [0.12-0.22] | anger: 0.17 [0.08-0.26] |
| Paranoia | negemo: 0.18 [0.08-0.29] | death: 0.15 [0.09-0.21] | shehe: 0.12 [0.08-0.16] |
| Planning | adj: 0.10 [0.04-0.17] | NS | NS |
| Relationship | verb: 0.18 [0.14-0.22] | health: 0.17 [0.13-0.21] | tentat: 0.16 [0.12-0.2] |
| Soldier | male: 0.13 [0.07-0.19] | shehe: 0.12 [0.08-0.16] | social: 0.11 [0.07-0.15] |
| Suicide | sad: 0.15 [0.11-0.19] | NS | NS |
| Surveillance | social: 0.13 [0.09-0.17] | shehe: 0.12 [0.08-0.16] | verb: 0.11 [0.07-0.16] |
| Threat | affect: 0.25 [0.14-0.36] | reward: 0.16 [0.06-0.25] | death: 0.15 [0.07-0.23] |
| Violence | Dic: 0.15 [0.10-0.19] | money: 0.12 [0.07-0.16] | Period: 0.08 [0.04-0.13] |
| Weaponry | death: 0.19 [0.07-0.31] | adverb: 0.09 [0.05-0.14] | adj: 0.09 [0.04-0.14] |

*Note.* [1]NS = not significant (i.e., fewer than three significantly correlating categories)

*Table 7.* Top 3 Strongest Pearson correlations German dictionary and LIWC

| Category | Strongest correlating categories | | |
|---|---|---|---|
| Deadline | achiev: 0.2 [0.09-0.31] | relativ: 0.13 [0.13-0.14] | drives: 0.12 [0.06-0.18] |
| Desperation | sad: 0.26 [0.25-0.27] | negemo: 0.25 [0.12-0.39] | affect: 0.21 [0.1-0.32] |
| Fixation | see: 0.12 [0.11-0.13] | insight: 0.09 [0.08-0.1] | percept: 0.08 [0.07-0.08] |
| Frustration | negemo: 0.28 [0.17-0.39] | risk: 0.23 [0.22-0.25] | sad: 0.19 [0.17-0.21] |
| God | relig: 0.25 [0.22-0.29] | interrog: 0.23 [0.15-0.31] | Tone: 0.1 [0.08-0.12] |
| Grievance | negemo: 0.44 [0.17-0.71] | sad: 0.32 [0.26-0.38] | affect: 0.24 [0.09-0.39] |
| Hate | affect: 0.24 [0.1-0.38] | sad: 0.16 [0.14-0.18] | differ: 0.07 [0.03-0.11] |
| Help | adj: 0.15 [0.13-0.17] | social: 0.14 [0.06-0.23] | affect: 0.13 [0.11-0.14] |
| Honour | posemo: 0.25 [0.16-0.34] | power: 0.19 [0.1-0.27] | affect: 0.18 [0.17-0.19] |
| Impostor | differ: 0.14 [0.1-0.17] | risk: 0.13 [0.12-0.14] | adj: 0.12 [0.11-0.13] |
| Jealousy | negemo: 0.21 [0.09-0.33] | risk: 0.12 [0.06-0.18] | affect: 0.11 [0.05-0.18] |
| Loneliness | i: 0.35 [0.27-0.42] | article: 0.28 [0.26-0.29] | ppron: 0.2 [0.09-0.3] |
| Murder | Dic: 0.12 [0.11-0.13] | sad: 0.11 [0.1-0.12] | health: 0.08 [0.05-0.12] |
| Paranoia | negemo: 0.23 [0.1-0.35] | anx: 0.2 [0.12-0.27] | health: 0.15 [0.12-0.17] |
| Planning | achiev: 0.12 [0.09-0.15] | Dic: 0.12 [0.11-0.13] | drives: 0.11 [0.11-0.12] |
| Relationship | we: 0.21 [0.09-0.33] | interrog: 0.17 [0.16-0.17] | Tone: 0.13 [0.11-0.14] |
| Soldier | social: 0.14 [0.12-0.15] | risk: 0.13 [0.12-0.14] | Clout: 0.13 [0.09-0.16] |
| Suicide | sad: 0.28 [0.27-0.29] | health: 0.26 [0.26-0.27] | affect: 0.16 [0.06-0.25] |
| Surveillance | see: 0.2 [0.18-0.21] | percept: 0.15 [0.12-0.19] | ppron: 0.15 [0.14-0.15] |
| Threat | ppron: 0.4 [0.36-0.45] | i: 0.38 [0.2-0.56] | pronoun: 0.24 [0.23-0.25] |
| Violence | ppron: 0.25 [0.17-0.33] | sad: 0.16 [0.15-0.17] | pronoun: 0.15 [0.11-0.18] |
| Weaponry | drives: 0.09 [0.08-0.1] | feel: 0.03 [0.01-0.04] | verb: 0.02 [0.01-0.03] |

*Table 8.* Top 3 strongest Pearson correlations Italian Dictionary and LIWC

| Category | Strongest correlating categories | | |
|---|---|---|---|
| Deadline | time: 0.16 [0.15-0.18] | past: 0.13 [0.13-0.14] | insight: 0.11 [0.1-0.11] |
| Desperation | sad: 0.55 [0.29-0.81] | negemo: 0.39 [0.22-0.56] | affect: 0.19 [0.13-0.25] |
| Fixation | feeling: 0.14 [0.06-0.21] | insight: 0.13 [0.13-0.14] | percept: 0.12 [0.08-0.16] |
| Frustration | negemo: 0.3 [0.2-0.39] | affect: 0.14 [0.13-0.15] | sentim: 0.14 [0.12-0.15] |
| God | relig: 0.25 [0.16-0.33] | metafis: 0.2 [0.12-0.29] | optimism: 0.04 [0.02-0.06] |
| Grievance | negemo: 0.19 [0.19-0.19] | sentim: 0.14 [0.09-0.18] | anx: 0.11 [0.07-0.16] |
| Hate | anger: 0.33 [0.31-0.36] | negemo: 0.26 [0.2-0.32] | sentim: 0.13 [0.12-0.14] |
| Help | affect: 0.31 [0.11-0.5] | sad: 0.24 [0.16-0.33] | dic: 0.13 [0.08-0.18] |
| Honour | posemo: 0.15 [0.06-0.23] | friend: 0.11 [0.11-0.12] | affect: 0.09 [0.04-0.14] |
| Impostor | dic: 0.07 [0.03-0.11] | excl: 0.06 [0.05-0.07] | cogmech: 0.02 [0.01-0.03] |
| Jealousy | negemo: 0.14 [0.11-0.16] | dic: 0.12 [0.1-0.15] | work: 0.12 [0.07-0.16] |
| Loneliness | self: 0.38 [0.29-0.47] | i: 0.32 [0.21-0.44] | sad: 0.26 [0.2-0.31] |
| Murder | prep: 0.2 [0.09-0.31] | death: 0.15 [0.09-0.22] | negemo: 0.15 [0.12-0.18] |
| Paranoia | anx: 0.21 [0.16-0.27] | negemo: 0.2 [0.15-0.24] | anger: 0.16 [0.13-0.19] |
| Planning | insight: 0.16 [0.15-0.16] | cogmech: 0.15 [0.14-0.16] | dic: 0.14 [0.06-0.22] |
| Relationship | posemo: 0.25 [0.1-0.41] | social: 0.17 [0.15-0.2] | i: 0.16 [0.16-0.17] |
| Soldier | cause: 0.29 [0.16-0.41] | prep: 0.22 [0.2-0.23] | cogmech: 0.18 [0.17-0.19] |
| Suicide | sad: 0.41 [0.17-0.64] | negemo: 0.34 [0.13-0.56] | dic: 0.22 [0.2-0.23] |
| Surveillance | see: 0.39 [0.3-0.49] | cause: 0.3 [0.21-0.39] | percept: 0.26 [0.14-0.39] |
| Threat | negemo: 0.16 [0.1-0.22] | inhib: 0.1 [0.1-0.1] | affect: 0.07 [0.05-0.09] |
| Violence | negemo: 0.28 [0.24-0.32] | anger: 0.25 [0.21-0.29] | affect: 0.15 [0.1-0.2] |
| Weaponry | negemo: 0.11 [0.1-0.12] | inhib: 0.1 [0.09-0.11] | dic: 0.07 [0.06-0.09] |

**Discussion and conclusion**

The current paper documented the translation process of the Grievance Dictionary and the psychometric evaluation of the translations. Overall, the translation into three languages (German, Dutch, Italian) resulted in internally reliable measures for most dictionary categories. In linguistic research, Cronbach's alpha scores obtained for internal reliability measurements of dictionaries are generally lower than in other research fields, such as classical survey research (Pennebaker et al., 2015), because word use is not repeated in the way that survey items are. The Dutch translation is closest to the reliability scores obtained in the English version of the Grievance Dictionary, followed by the German version. The range of reliability scores is lower for all translated versions, but this is not unexpected given that the terms were translated from English and not generated in a bottom-up manner (i.e., developing new dictionaries from scratch). We accept this as a trade-off of the chosen approach. In Italian, however, there are several categories that showed considerably low reliability, suggesting the words in the category might represent another concept or that the category itself consists of multiple dimensions that are oversimplified here into one single category. As a consequence, we urge users to apply caution when measuring the Grievance categories fixation, frustration, jealousy, and loneliness in the Italian translation. Further research is needed to understand whether and how these categories need to be adjusted in Italian. Furthermore, the correlations between translated Grievance dictionary and LIWC categories were in line with results obtained in the English version. Most of the correlations obtained seem to be among psychologically related concepts, with the LIWC category often functioning as an 'umbrella' for the Grievance dictionary categories it correlates with (e.g., LIWC 'negative emotion' correlates with 'jealousy', 'murder' and 'paranoia' from the Dutch dictionary).

Our hope is that the procedure documented here inspires other researchers to translate the Grievance dictionary into more languages and make these publicly available. Some limitations of our translation approach need to be considered, which could potentially be addressed in future endeavours. First, we relied on Google Translate for initial term translations, followed by a stepwise procedure of quality controls and corrections by human annotators. While previous research has shown that a two-step procedure of automated translations and subsequent human correction results in sufficiently similar dictionary translations as a fully human approach (Boot et al., 2017), it is not known for our translations how they compare to a human-only approach, or to translations derived from tools other than Google Translate. Although manual translation of several thousand terms is a significant task, an approach where the context for each term is taken into account (i.e., the overarching category and its meaning in threat assessment), may be particularly promising. Second, it may be beneficial to include a larger number of corpora for the psychometric analysis. At present, the calculations were done on two corpora per language (compared to four corpora in the English language version). Although the chosen corpora represented different contexts (e.g., comments, book reviews, tweets), it was more difficult to obtain appropriate non-English corpora for the purpose of this study. The difficulties associated with finding appropriate corpora in Dutch, Italian, and German are also the reason we were not able to perform a full validation of the dictionary as was done in the English version, which included statistical comparisons and machine learning classifications of grievance-fuelled texts.

The original Grievance dictionary was developed for the purpose of threat assessment, specifically, analysing language use in the context of grievance-fuelled violence. Results showed that the English version dictionary could distinguish between language use of extremist, violent, and non-violent populations, respectively. A host of other papers have shown the applicability of the dictionary for related purposes as well as in other domains. The next step in validating the German, Dutch, and Italian versions of the dictionary would be to apply the dictionary to data in each respective language, enabling new research in threat assessment, extremism, and violence in languages other than English. We encourage the research community to build on this initiative and further explore its potential.

**CRediT statement**
IV led conceptualization, analysis, and writing; BK contributed to conceptualization, annotation, and writing; MM contributed to annotation and review; JF contributed to annotation, review and editing. The authors would like to thank the additional annotators for their help in translating the dictionary.

# References

Ai, L., Chen, Y.-W., Yu, Y., Kweon, S., Hirschberg, J., & Levitan, S. I. (2024). *What Makes A Video Radicalizing? Identifying Sources of Influence in QAnon Videos* (arXiv:2404.14616). arXiv. https://doi.org/10.48550/arXiv.2404.14616

Allchorn, W., Dafnos, A., & Gentile, F. (2022). The Role of Violent Conspiratorial Narratives in Violent and Non-Violent Extreme Right Manifestos Online, 2015-2020. *Global Research Network on Terrorism and Technology (GNET)*.

Assenmacher, D., Niemann, M., Müller, K., Seiler, M., Riehle, D. M., & Trautmann, H. (2021, August 29). *RP-Mod & RP-Crowd: Moderator- and Crowd-Annotated German News Comment Datasets*. Thirty-fifth Conference on Neural Information Processing Systems Datasets and Benchmarks Track (Round 2). https://openreview.net/forum?id=NfTU-wN8Uo

Bianchi, F., Nozza, D., & Hovy, D. (2021). FEEL-IT: Emotion and Sentiment Classification for the Italian Language. In O. De Clercq, A. Balahur, J. Sedoc, V. Barriere, S. Tafreshi, S. Buechel, & V. Hoste (Eds.), *Proceedings of the Eleventh Workshop on Computational Approaches to Subjectivity, Sentiment and Social Media Analysis* (pp. 76–83). Association for Computational Linguistics. https://aclanthology.org/2021.wassa-1.8/

Boot, P., Zijlstra, H., & Geenen, R. (2017). The Dutch translation of the Linguistic Inquiry and Word Count (LIWC) 2007 dictionary. *Dutch Journal of Applied Linguistics*, *6*(1), 65–76. https://doi.org/10.1075/dujal.6.1.04boo

Burgh, B. van der, & Verberne, S. (2019). *The merits of Universal Language Model Fine-tuning for Small Datasets—A case with Dutch book reviews* (arXiv:1910.00896). arXiv. https://doi.org/10.48550/arXiv.1910.00896

Cerone, L. (2023). *Reddit Italy Coffee Dataset*. https://www.kaggle.com/datasets/gigggi/reddit-italy-coffee-dataset

Edmondson, M., Muschelli, J., Richardson, N., & Gustavsen, J. (2020). *googleLanguageR: Call Google's "Natural Language" API, "Cloud Translation" API, "Cloud Speech" API and "Cloud Text-to-Speech" API* (Version 0.3.0) [Computer software]. https://cran.r-project.org/web/packages/googleLanguageR/index.html

Gwet, K. L. (2008). Computing inter-rater reliability and its variance in the presence of high agreement. *British Journal of Mathematical and Statistical Psychology*, *61*(1), 29–48. https://doi.org/10.1348/000711006X126600

Habib, H., Srinivasan, P., & Nithyanand, R. (2022). Making a Radical Misogynist: How Online Social Engagement with the Manosphere Influences Traits of Radicalization. *Proc. ACM Hum.-Comput. Interact.*, *6*(CSCW2), 450:1-450:28. https://doi.org/10.1145/3555551

Klonek, F. E., Gerpott, F. H., & Handke, L. (2022). When Groups of Different Sizes Collide: Effects of Targeted Verbal Aggression on Intragroup Functioning. *Group & Organization Management*, 10596011221134426. https://doi.org/10.1177/10596011221134426

Lanning, K., Wetherell, G., Warfel, E. A., & Boyd, R. L. (2021). Changing channels? A comparison of Fox and MSNBC in 2012, 2016, and 2020. *Analyses of Social Issues and Public Policy*, *21*(1), 149–174. https://doi.org/10.1111/asap.12265

Luyckx, K., & Daelemans, W. (2008). Personae: A Corpus for Author and Personality Prediction from Text. In N. Calzolari, K. Choukri, B. Maegaard, J. Mariani, J. Odijk, S. Piperidis, & D. Tapias (Eds.), *Proceedings of the Sixth International Conference on Language Resources and Evaluation (LREC`08)*. European Language Resources Association (ELRA). https://aclanthology.org/L08-1030/

Meier, T., Boyd, R. L., Pennebaker, J. W., Mehl, M. R., Martin, M., Wolf, M., & Horn, A. B. (2019). *"LIWC auf Deutsch": The Development, Psychometrics, and Introduction of DE-LIWC2015*. https://doi.org/10.31234/osf.io/uq8zt

Pennebaker, J. W., Boyd, R. L., Jordan, K., & Blackburn, K. (2015). *The Development and Psychometric Properties of LIWC2015*. The University of Texas at Austin. https://repositories.lib.utexas.edu/handle/2152/31333

Schabus, D., Skowron, M., & Trapp, M. (2017). One Million Posts: A Data Set of German Online Discussions. *Proceedings of the 40th International ACM SIGIR Conference on Research and Development in Information Retrieval*, 1241–1244. https://doi.org/10.1145/3077136.3080711

van der Vegt, I., Mozes, M., Kleinberg, B., & Gill, P. (2021). The Grievance Dictionary: Understanding threatening language use. *Behavior Research Methods*. https://doi.org/10.3758/s13428-021-01536-2

Wongpakaran, N., Wongpakaran, T., Wedding, D., & Gwet, K. L. (2013). A comparison of Cohen's Kappa and Gwet's AC1 when calculating inter-rater reliability coefficients: A study conducted with personality disorder samples. *BMC Medical Research Methodology*, *13*(1), 61. https://doi.org/10.1186/1471-2288-13-61